\pdfoutput=1

\documentclass[11pt]{article}

\usepackage[preprint]{acl}
\usepackage{url}
\usepackage{hyperref}
\usepackage{times}
\usepackage{latexsym}

\usepackage[T1]{fontenc}

\usepackage[utf8]{inputenc}

\usepackage{microtype}

\usepackage{inconsolata}

\usepackage{graphicx}

\usepackage{xcolor}

\usepackage{enumitem}

\DeclareRobustCommand{\legendsquare}[1]{%
  \textcolor{#1}{\rule{1ex}{1ex}}%
}

\definecolor{darkgreen}{RGB}{0,90,0}
\definecolor{lightred}{RGB}{255,70,70}
\title{Beyond Silent Letters: Amplifying LLMs in\\Emotion Recognition with Vocal Nuances}

  \author {
    Zehui Wu, \textsuperscript{\rm 1}
    Ziwei Gong, \textsuperscript{\rm 1}
    Lin Ai, \textsuperscript{\rm 1}
    Pengyuan Shi,\textsuperscript{\rm 1}
    Kaan Donbekci, \textsuperscript{\rm 1}
    Julia Hirschberg \textsuperscript{\rm 1}\\
  Department of Computer Science\\Columbia University\\
  \texttt{\{zw2804, zg2272, la2734, ps3391, kd2939\}@columbia.edu} \\
  \texttt{julia@cs.columbia.edu}
}

\begin{document}
\maketitle
\begin{abstract}

Emotion recognition in speech is a challenging multimodal task that requires understanding both verbal content and vocal nuances. This paper introduces a novel approach to emotion detection using Large Language Models (LLMs), which have demonstrated exceptional capabilities in natural language understanding. To overcome the inherent limitation of LLMs in processing audio inputs, we propose \textit{SpeechCueLLM}, a method that translates speech characteristics into natural language descriptions, allowing LLMs to perform multimodal emotion analysis via text prompts without any architectural changes. Our method is minimal yet impactful, outperforming baseline models that require structural modifications. We evaluate \textit{SpeechCueLLM} on two datasets: IEMOCAP and MELD, showing significant improvements in emotion recognition accuracy, particularly for high-quality audio data. We also explore the effectiveness of various feature representations and fine-tuning strategies for different LLMs. Our experiments reveal that incorporating speech descriptions leads to an improvement of nearly 10 points in the F1 score under the zero-shot setting and over 2.5 points under the LoRA setting on the IEMOCAP dataset. The source code will be made available on GitHub.

\end{abstract}

\section{Introduction}
Emotion detection in speech is a crucial aspect of human-computer interaction, with applications spanning from customer service to mental health monitoring \cite{van2023emotion,9544838}. Recent research has demonstrated that Large Language Models (LLMs) possess a form of emotional intelligence, capable of interpreting emotional stimuli in text \cite{li2023largelanguagemodelsunderstand}. Advances in Emotion Recognition in Conversation (ERC) have utilized LLMs to improve performance and generalizability \cite{lei2024instructercreformingemotionrecognition, xue2024biosercintegratingbiographyspeakers, fu2024ckercjointlarge}. 

However, despite the remarkable progress of LLMs in handling textual data, they remain fundamentally limited in processing audio inputs. This limitation constrains their utility in multimodal emotion recognition tasks, which require the integration of both verbal content and vocal nuances. While recent studies have explored incorporating speech features into LLM-based systems \cite{xu2023secapspeechemotioncaptioning, zhang2024llmsmeetsacousticlandmarks, Gong_2023, 10448316}, these approaches often introduce additional model parameters and pre-training stages, increasing complexity and computational overhead.

In this paper, we address these challenges by proposing \textit{SpeechCueLLM}, a novel method that translates speech characteristics into natural language descriptions, enabling LLMs to process multimodal data through text prompts without requiring any architectural changes. This approach bridges the gap between audio and text modalities, allowing LLMs to conduct multimodal emotion analysis. \textit{SpeechCueLLM} sets itself apart from existing speech-integrated LLM approaches in three key aspects: \textbf{(a) Direct Integration}: Our method allows seamless integration of speech characteristics without adding new neural components (e.g., audio encoders or projection layers), thus simplifying the training process. \textbf{(b) End-to-End Fine-tuning}: We fine-tune the LLMs directly on inputs that combine textual data with speech descriptions, eliminating the need for weak label generation or additional training stages. \textbf{(c) Flexibility and Generalization}: By translating speech features into natural language descriptions, \textit{SpeechCueLLM} provides a flexible, plug-in solution for incorporating speech information into LLM-based systems without modifying the LLM architecture.

Our main contributions are as follows:
\begin{itemize}[nosep,topsep=1pt]
    \item[1.] We introduce \textit{SpeechCueLLM}, a method that enables multimodal analysis by integrating speech characteristics into text prompts, facilitating emotion recognition without architectural changes to the LLMs. \textit{SpeechCueLLM} outperforms traditional approaches that incorporate speech encoders for embedding audio features into LLMs.
    \item[2.] We present key findings on the effectiveness and limitations of this method, showing that incorporating speech descriptions enhances emotion recognition accuracy across multiple LLMs while offering insights into the role of audio quality and the advantages of prompt-based techniques.
\end{itemize}

This study aims to showcase the potential of \textit{SpeechCueLLM} to enhance the emotion recognition capabilities of LLMs while providing insights into the factors driving its effectiveness. We evaluate \textit{SpeechCueLLM} on two widely used datasets in ERC: IEMOCAP \cite{Busso2008IEMOCAP} and MELD \cite{poria2019meldmultimodalmultipartydataset} (Section \ref{sec:datasets}), achieving state-of-the-art performance. We compare our method with projection-based approaches that incorporate a separate speech encoder and show that \textit{SpeechCueLLM} surpasses these baseline methods (Section \ref{sec:baselines}). Additionally, we explore the impact of different speech feature representations (Section \ref{sec:features}) and examine the fine-tuning performance of various LLMs (Section \ref{sec:llms}).

\section{Related Work}

\subsection{Multimodal Emotion Models}

 Exsiting multimodal methods have been proposed to combine audio and text modalities for emotion recognition \cite{zadeh2017tensorfusionnetworkmultimodal, wu-etal-2024-multimodal, meng2024revisitingmultimodalemotionrecognition, tsai2019multimodaltransformerunalignedmultimodal, 9928357, 10109845}. Many of these works utilize transformer encoder structures to fuse multimodal signals through cross-modality attention mechanisms \cite{tsai2019multimodaltransformerunalignedmultimodal}. Additionally, some models employ variations of Graph Neural Networks (GNNs) to capture multimodal semantic features and enhance emotion recognition accuracy \cite{meng2024revisitingmultimodalemotionrecognition}. These methods focus on model structure that captures and learns from multimodal signals holistically.

\subsection{LLMs for Emotion Recognition}

Recent advancements in ERC have leveraged LLMs to enhance performance and generalizability. In our work, we build upon the InstructERC framework \cite{lei2024instructercreformingemotionrecognition}, which reformulates emotion recognition as a generative task using LLMs. While InstructERC introduced retrieval templates and emotion alignment tasks, recent extensions such as BiosERC \cite{xue2024biosercintegratingbiographyspeakers}, CKERC \cite{fu2024ckercjointlarge}, and DialogueLLM \cite{zhang2024dialoguellmcontextemotionknowledgetuned} have incorporated additional contextual information like biographical data, commonsense knowledge, and visual descriptions.

Our approach distinguishes itself by focusing on integrating speech characteristics into the templates as natural language descriptions, bridging the gap between audio and text modalities in LLM-based emotion recognition systems.

\subsection{Speech Incorporation in LLMs}
Recent works have explored integrating speech features into LLM-based systems for various downstream tasks. SECap \cite{xu2023secapspeechemotioncaptioning} uses LLaMA as a text decoder to generate speech emotion captions, relying on an additional audio encoder and bridging network. Similarly, \cite{zhang2024llmsmeetsacousticlandmarks} incorporates acoustic landmark tokens for depression detection and requires extra parameters and pre-training.

LanSER \cite{Gong_2023} infers weak emotion labels from speech transcripts to use unlabeled data, while \cite{10448316} focuses on emotional speech annotation, combining conversation context with acoustic feature descriptors. While this approach shares our interest in combining textual and acoustic information in the LLM inputs, it primarily uses LLMs for weak label generation rather than direct emotion recognition training.

Recent researchhas also explored the application of LLMs in Automatic Speech Recognition (ASR)\cite{bai2024seedasrunderstandingdiversespeech,baskar2024speechprefixtuningrnntloss,ma2024embarrassinglysimpleapproachllm} , demonstrating the versatility of LLMs in speech-processing tasks. Further extending the capabilities of LLMs in speech processing, AudioChatLlama \cite{fathullah-etal-2024-audiochatllama} and Audio-LLM \cite{10.1007/978-981-97-4399-5_13} presents end-to-end model that integrates general-purpose speech abilities into the Llama-2 model.

\textit{SpeechCueLLM} differs from these approaches by processing audio-derived information without requiring architectural modifications. This method is highly compatible with existing LLM infrastructures, facilitating easy implementation and integration into current systems.


\section{Methodology}
\begin{figure}[t]
  \includegraphics[width=\columnwidth]{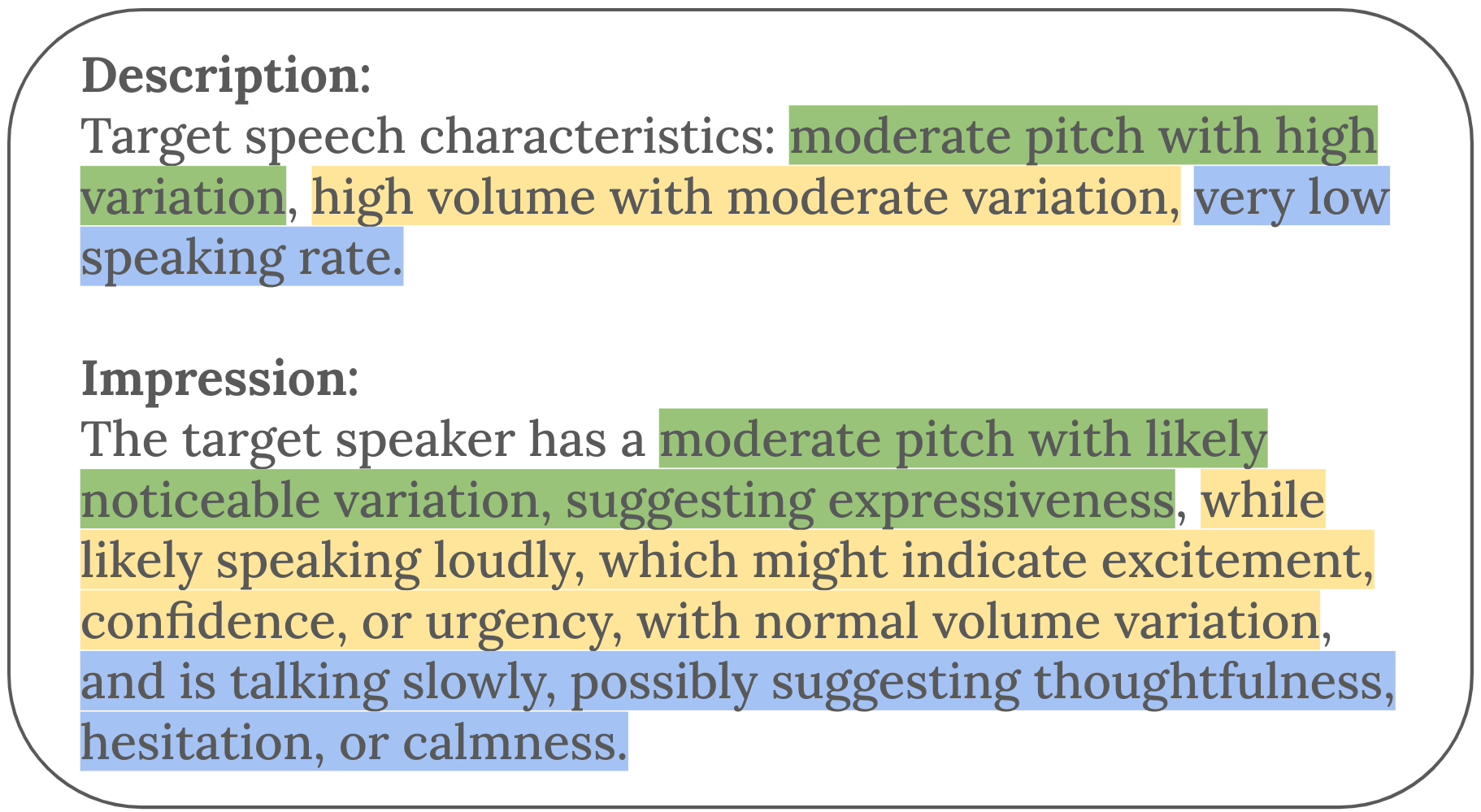}
  \caption{Example of speech characteristic description (top) and derived impression (bottom). Each color represents the same set of features (\legendsquare{green} pitch, \legendsquare{yellow} volume, and \legendsquare{blue} speaking rate).}
  \label{fig:des}
\end{figure}

\textit{SpeechCueLLM} integrates descriptions of speech characteristics in natural language into the prompt for LLMs, as illustrated in Figure \ref{fig:des}. Traditional LLMs, when limited to textual inputs, omit critical emotional cues that are embedded in audio signals. This omission can result in a failure to capture important aspects of emotion, as the same sentence can express different emotions based on variations in pitch, volume, intonation, and other vocal features. By translating audio signals into natural language descriptions, \textit{SpeechCueLLM} enables LLMs to perform more accurate and nuanced multimodal emotion recognition.

\subsection{Audio Features}
In our approach to emotion detection, we leverage five well-established and intuitive audio features that convey emotional cues \cite{koolagudi2012emotion,7226047, 7155930}. These features are designed to be easily interpreted by both humans and LLMs, capturing key aspects of speech that reflect emotional information. 

The first feature, \textit{volume}, is represented by two sub-features: \textit{average volume}, which indicates the overall loudness of the speech, and \textit{volume variation}, which captures dynamic changes in intensity. These features allow us to detect emotional nuances based on loudness. 

\textit{Pitch}, another critical indicator of emotion, is also divided into two sub-features: \textit{average pitch}, reflecting the general tone of the speaker's voice, and \textit{pitch variation}, which highlights modulation and changes in pitch, providing deeper insights into the speaker's emotional state.

The fifth feature, \textit{speaking rate}, measures the speed of speech, offering valuable information about the speaker’s emotional urgency or level of excitement. 

By focusing on these fundamental yet comprehensive audio characteristics, we ensure that the extracted features are rich in emotional content while remaining interpretable by LLMs when translated into natural language descriptions.

\subsubsection{Audio Feature Processing}
Our approach to audio feature processing transforms raw numerical features into categorical representations to enhance interpretability and facilitate natural language descriptions. This process is composed of four key steps:

\begin{enumerate}
    \item \textbf{Threshold Calculation}: We compute thresholds for each audio feature (such as average volume, pitch variation, and speaking rate) using quantile-based segmentation. Depending on the desired granularity, the feature space is divided into 3, 4, 5, or 6 classes, with thresholds determined by appropriate quantiles. Detailed quantile splits can be found in Table \ref{table:quantile_splits} in Appendix \ref{sec:appendix_1}.
    
    \item \textbf{Speaker-Specific Normalization}: To account for individual differences in speech patterns, we calculate speaker-specific thresholds, falling back to overall thresholds for less frequent speakers. This allows for more accurate categorization of speech features relative to each speaker’s typical speaking style.
    
    \item \textbf{Categorization}: Based on the calculated thresholds, each numerical feature is categorized. For example, in a 5-class system, a feature value may be classified as "very low," "low," "medium," "high," or "very high." This categorization provides a more intuitive, human-readable representation of the audio features.
    
    \item \textbf{Feature-Specific Descriptions}: For each key audio feature (volume, pitch, speaking rate), we generate descriptive phrases corresponding to their categorical values. For instance, "high volume with moderate variation" or "low pitch with high variation" offers a more accessible and interpretable description of the audio features.
\end{enumerate}

\subsubsection{Audio Impression}

To make the audio features more accessible and meaningful for both human interpretation and LLM processing, we generate hard-coded natural language impressions based on the categorized features. These impressions go beyond simple descriptions, offering interpretative insights that suggest potential emotional or psychological states associated with the observed speech patterns. The feature-impression mapping is manually curated based on established relationships between prosodic features and emotional states \cite{mozziconacci2002prosody,frick1985communicating}.

For example, "speaking loudly with significant volume changes" might be interpreted as indicating "excitement, confidence, or urgency." The full mapping of features to impressions can be found in Table \ref{table:Feature-Impression Mapping} in Appendix \ref{sec:appendix_1}.

To account for the inherent uncertainty in interpreting speech patterns, we incorporate hedge words (e.g., "likely," "may") depending on how close a feature is to its categorical threshold. This nuanced approach prevents overconfident interpretations, particularly in borderline cases.

The final impression synthesizes information from pitch, volume, and speaking rate into a coherent, flowing sentence. This integration provides a rich description of the speaker's vocal characteristics and their potential emotional implications.

\subsection{\textit{SpeechCueLLM} Prompt Template}

\begin{figure}[t]
\centering
  \includegraphics[width=\columnwidth]{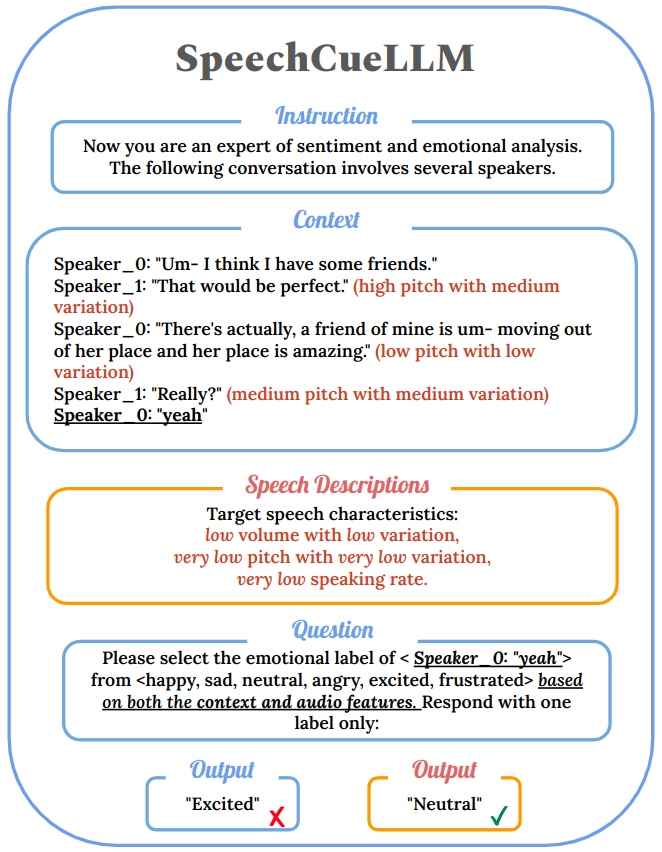}
  \caption{\textit{SpeechCueLLM} Prompt Template for Emotion Detection: the last bold sentence with an underline represents the target utterance. The orange section highlights the outputs with added speech descriptions. This structured template integrates both textual context and speech characteristics to guide the LLM in performing multimodal emotion analysis.}
  \label{fig:speechcuellm}
\end{figure}

Our methodology employs a comprehensive prompt template designed to enhance emotion detection performance. As shown in Figure \ref{fig:speechcuellm}, the template is structured around four key components:

\begin{enumerate}[nosep,topsep=1pt]
  \item \textbf{Instruction}: This component sets the stage by positioning the LLM as an expert in sentiment and emotion analysis, priming it for the task.
  
  \item \textbf{Context}: Providing conversational background is essential for helping the LLM understand the broader situation and interpersonal dynamics. Additionally, speech features are appended to the last three utterances, offering supplementary speech context. More details on the added speech context are discussed in Appendix \ref{sec:appendix_2}.
  
  \item \textbf{Speech Descriptions}: This innovative element translates audio signals into natural language descriptions of key speech characteristics like volume, pitch, and speaking rate. By including these paralinguistic details, we bridge the gap between audio and text, enabling the LLM to account for speech features that are crucial for accurate emotion detection.
  
  \item \textbf{Question}: This final component asks the LLM to select an emotional label for a target utterance from a predefined set of options. It explicitly directs the model to consider both the contextual background and the described audio features in making its decision.
\end{enumerate}

Our empirical analysis shows that straightforward prompts with simple instructions yield optimal results. This structured approach ensures that the LLM can access both textual and audio-derived information, leading to a more nuanced and accurate emotion detection process. By incorporating speech characteristics into the prompt, we allow the LLM to perform multimodal analysis without requiring direct audio input processing capabilities.

\section{Experiments}

\subsection{Datasets}
\label{sec:datasets}
We utilize two widely recognized datasets in the field of multimodal emotion recognition:

\textbf{IEMOCAP (Interactive Emotional Dyadic Motion Capture)} \cite{Busso2008IEMOCAP}: IEMOCAP consists of dyadic conversations between actors, providing a rich set of emotional expressions in a controlled environment. It contains approximately 12 hours of audiovisual data from 10 actors, divided into 5 sessions. Each utterance is annotated with one of nine emotion labels: anger, happiness, excitement, sadness, frustration, fear, surprise, neutral, and other. IEMOCAP is particularly valued for its high-quality audio recordings and motion capture data, making it an ideal dataset for multimodal emotion analysis.

\textbf{MELD (Multimodal EmotionLines Dataset)} \cite{poria2019meldmultimodalmultipartydataset}: MELD is derived from the TV series \textit{Friends}, making it particularly useful for analyzing emotions in natural conversational settings. It contains 13,708 utterances from 1,433 dialogues, each annotated with one of seven emotion categories: anger, disgust, fear, joy, neutral, sadness, and surprise. 

In our experiments, we focus on the audio modality from both datasets. For the MELD dataset, we implement speaker-based standardization of audio features to account for individual vocal characteristics. In the IEMOCAP dataset, we apply gender-based standardization to account for physiological differences that can affect vocal features. Additionally, we confine our analysis of IEMOCAP to six emotion categories (anger, happiness, excitement, sadness, frustration, and neutral), following common practices in the field \cite{gong-etal-2024-mapping}.

Our use of both the MELD and IEMOCAP datasets is designed to evaluate our approach across diverse audio environments. MELD, originating from a television show, presents more chaotic and naturalistic scenarios, while IEMOCAP offers a more controlled, studio-based environment. This dual-dataset approach enables a comprehensive assessment of our emotion detection method under varied recording conditions.

\subsection{Experiment Settings}

\subsubsection{\textit{SpeechCueLLM} Training Settings}
For our experiments, we focus on open-sourced LLMs and include comparisons with the best-performing proprietary LLMs. We input the structured prompts into the LLMs and apply LoRA \cite{hu2021loralowrankadaptationlarge}, an efficient low-parameter fine-tuning method that minimizes overfitting. This method trains the model to output a single emotion label based on the prompt. To further validate our approach, we also conduct experiments using only speech features in the prompts, demonstrating the effectiveness of these simple features. 

We experiment with fine-tuning various LLMs, including LLaMA-2, multiple versions of LLaMA-3, and phi-3-medium, to evaluate the generalizability of our approach. For training, we use a learning rate of 3e-4 for IEMOCAP and 2e-4 for MELD, with a linear warm-up scheduler. Additionally, we test the zero-shot performance of Claude Sonnet 3.5 for comparison. Results are averaged over three independent runs to ensure robustness.

\subsubsection{Baselines: LLMs with Speech Encoder}
To benchmark our approach, we implement a projection-based method that leverages speech features by training additional parameters. In this setup, audio features are encoded using a pre-trained Automatic Speech Recognition (ASR) model and projected into the LLM's embedding space, as illustrated in Figure \ref{fig:projector}. This approach is inspired by SLAM-ASR \cite{ma2024embarrassinglysimpleapproachllm}, which combines off-the-shelf speech encoders with LLMs using a trainable linear projector, achieving state-of-the-art performance on the Librispeech benchmark \cite{7178964}. 

For consistency, the text portion of the input follows the same prompt format as in our main approach, but without the inclusion of speech-derived features. The encoded audio tokens are placed before the text tokens in the input sequence.

We use the Data2Vec model \cite{baevski2022data2vecgeneralframeworkselfsupervised} as the speech encoder. To prevent overfitting, we apply a downsampler to reduce the feature sequence length from 50 tokens per second to 10 tokens per second. For projecting the speech features into the LLM embedding space, we employ two types of projectors: 
\begin{enumerate}[nosep,topsep=1pt]
    \item A straightforward feed-forward neural network.
    \item A Q-former \cite{li2023blip2bootstrappinglanguageimagepretraining}, a query-based transformer model that uses learnable queries to attend to speech features, producing a condensed and structured representation of key information from the audio.
\end{enumerate}

Both projectors transform the encoded audio features into a format compatible with the LLM's token embeddings. As with the main approach, we fine-tune the LLM backbone using LoRA for all baseline methods.

\subsection{Results}
\label{sec:baselines}

\subsubsection{Comparison with the Baselines}

\begin{table*}
    \centering
    \begin{tabular}{lcccc}
    \hline
    \textbf{Approach} &\textbf{Trainable parameters} & \textbf{IEMOCAP F1 Score} & \textbf{MELD F1 Score} \\
    \hline
    Speech-Encoder-Only & 93.2M & 52.312 \scriptsize{$\pm$1.7} \small{(48.263)}  & 47.921 \scriptsize{$\pm$0.27} \small{(28.413)} \\
    \hline
    LLM \footnotesize{(w/o Speech Features)} & 12.6M & 70.111 \scriptsize{$\pm$0.36}  & 67.44 \scriptsize{$\pm$0.26}  \\
    \small{+FeedForward\textsubscript{Projector} +SE\textsubscript{Freeze}}      & 28.9M & 70.221\scriptsize{$\pm$0.2}  & 67.402\scriptsize{$\pm$0.27}  \\
    \small{+FeedForward\textsubscript{Projector} +SE\textsubscript{Unfreeze}}   & 122.1M  & 70.635\scriptsize{$\pm$0.74}  & 66.156\scriptsize{$\pm$0.83}  \\
    \small{+Q-Former\textsubscript{Projector} +SE\textsubscript{Freeze}}  & 71.4M & 70.235\scriptsize{$\pm$0.66}  & 67.194\scriptsize{$\pm$0.15}  \\
    \hline
    \textit{SpeechCueLLM} \footnotesize{(Ours)} & 12.6M & \textbf{72.596} \scriptsize{$\pm$0.26}  & \textbf{67.604} \scriptsize{$\pm$0.38}  \\
    \hline
    \end{tabular}
\caption{Performance (weighted F1) comparison between \textit{SpeechCueLLM} and the baselines. The top section includes results using only the speech encoder. (The number in parenthesis is the average F1 score.) The middle section includes results using only the LLM without speech features and varitions of the projection-based model. The bottom is our \textit{SpeechCueLLM} results.}
\label{tab: encoder}
\end{table*}

Our results show that \textit{SpeechCueLLM} outperforms baselines that embed audio features using speech encoders, achieving state-of-the-art (SOTA) performance, as shown in Table \ref{tab: encoder}.

We evaluate the effectiveness of speech features fine-tuned with the speech encoder. When used standalone, the audio encoder achieves a weighted F1 score of around 50 on both datasets. However, for MELD, the model tends to overfit the majority class, resulting in a reduced average F1 score of 28.413, while IEMOCAP achieves 48.263. This disparity in performance between the two datasets can be attributed to differences in audio quality and recording conditions. IEMOCAP, recorded in a controlled studio environment, provides clean and consistent audio, facilitating more accurate speech feature extraction. In contrast, MELD, derived from a TV show, \textit{Friends}, presents more challenging audio conditions. The presence of background laughter, varying audio quality, overlapping speakers, and short utterance durations (averaging 2-3 seconds) likely contribute to less reliable speech feature extraction, thereby limiting the benefits of speech features.

We conduct several experiments with different variations, including alternative projectors and either freezing or training the speech encoder. The results, shown in Table \ref{tab: encoder}, indicate slight improvements for IEMOCAP over the text-only model across all variations. Unfreezing the speech encoder leads to a marginally higher average score of 70.635\%, though at the cost of increased performance variability. Notably, the Q-former, despite having significantly more parameters than the feed-forward layers, delivers comparable results. In contrast, for MELD, all variations perform worse than the text-only model, likely due to the poor audio quality.

The performance gains in IEMOCAP from the projection-based approach (0.1-0.5) are notably smaller than those achieved using the prompt-based method (2.4). The prompt-based approach consistently outperforms the direct integration of speech encoder features into the LLM embedding space for several reasons. First, it preserves contextual coherence by representing speech characteristics as natural language descriptions, allowing the LLM to leverage its strengths in language processing. In contrast, inserting speech embeddings into the LLM’s embedding space can cause discrepancies in the distribution, introducing training complexity. This method may benefit from additional pre-training to align embeddings from the two modalities using large amount of text-audio paired data, as supported by previous research \cite{xu2023secapspeechemotioncaptioning, li2019visualbertsimpleperformantbaseline, arjmand2021teaseltransformerbasedspeechprefixedlanguage}. By sidestepping these multimodal alignment challenges, the prompt-based method delivers more consistent and reliable results without requiring extra tunings.

Furthermore, the prompt-based method enhances generalization by converting abstract speech features into textual descriptions, making it adaptable to various tasks and languages. This approach also offers greater interpretability, as the model explicitly receives human-readable descriptions of vocal nuances, unlike embedding modifications that obscure the model’s processing of audio features. Finally, the prompt-based method is computationally efficient, avoiding the need for complex architectural modifications or additional training, making it a lightweight yet highly effective solution.

\subsubsection{Comparison with the SOTAs}

\begin{table*}[h!]
\centering
\begin{tabular}{l|c|c|c}
\hline
\textbf{Method} & \textbf{Modality} & \textbf{IEMOCAP F1 Score} & \textbf{MELD F1 Score} \\ \hline
SDT \cite{SDT} & T, A, V & \textbf{74.08} & 66.6 \\ 
CFN-ESA \cite{CFNESA} & T, A, V & 71.04 & 66.70 \\ 
GS-MCC \cite{GSMCC} & T, A, V & 73.9 & 69.0 \\ 
ELR-GNN \cite{ELRGNN} & T, A, V & 70.9 & \textbf{69.9} \\ 
MPT-HCL \cite{MPTHCL} & T, A, V & 72.51 & 65.02 \\ 
TELME \cite{TELME} & T, A, V & 70.48 & 67.37 \\ 
Mamba-like Model \cite{MambaModel} & T, A, V & 73.3 & 67.6 \\ \hline
SpeechCueLLM (ours) & T, A & 72.6 & 67.6 \\ \hline
\end{tabular}
\caption{Performance comparison of methods on the IEMOCAP and MELD datasets. Modality: T (Text), A (Audio), V (Video).}
\label{table:sota}
\end{table*}

We collected results from the best-performing multimodal models in the recent two years for comparison. As shown in Table \ref{table:sota} Despite using only two modalities (Text and Audio), SpeechCueLLM's performance is on par with SOTA models that utilize three modalities (Text, Audio, and Video) for both datasets. Our LLM-based approach also provides more interpretability for both modalities. Moreover, SpeechCueLLM requires only 12.8 million trainable parameters, which is less than two linear projection layers in the projection-based baseline, highlighting its efficiency.

\subsection{Ablation Study and Discussion}
\subsubsection{How Effective Are Various Speech Features?}
\label{sec:features}
\begin{table*}
  \centering
  \begin{tabular}{l|cccc}
    \hline
    \textbf{Method} & \textbf{Text-only} & \textbf{+ speech des} & \textbf{+ speech imp} & \textbf{+ speech des with context}\\
    \hline
    \textbf{IEMOCAP} & 70.111 \scriptsize{$\pm$0.36} & 72.021 \scriptsize{$\pm$0.54} \textcolor{darkgreen}{\scriptsize{(+1.910)}} & 71.542 \scriptsize{$\pm$1.12} \textcolor{darkgreen}{\scriptsize{(+1.431)}} & \textbf{72.596} \scriptsize{$\pm$0.26} \textcolor{darkgreen}{\scriptsize{(+2.485)}}\\
    \textbf{MELD} & 67.44 \scriptsize{$\pm$0.26} & 67.074 \scriptsize{$\pm$0.49} \textcolor{lightred}{\scriptsize{(-0.366)}} & \textbf{67.604} \scriptsize{$\pm$0.38} \textcolor{darkgreen}{\scriptsize{(+0.164)}} & 67.09 \scriptsize{$\pm$0.57} \textcolor{lightred}{\scriptsize{(-0.35)}}\\
    \hline
  \end{tabular}
  \caption{\textit{SpeechCueLLM} performances (weighted F1 score in \%) on IEMOCAP and MELD datasets using LoRA fine-tuning on the LLaMA-2-7B-base model. Results are averaged over three independent runs, reported as mean $\pm$ standard deviation. Dark green values in parentheses show the improvement over the Text-only method.}
  \label{tab:main_results}
\end{table*}

The results in Table \ref{tab:main_results} demonstrate that incorporating speech features into text input improves performance across both datasets, though to varying degrees. For the IEMOCAP dataset, we observe a significant improvement of nearly 2 percentage points when adding speech descriptions to the text input (from 70.111\% to 72.021\%). By further enriching the input with additional speech features for context, the model improves to 72.596\% (detailed context experiments are available in Appendix \ref{sec:appendix_2}). This notable gain highlights the importance of integrating acoustic information in emotion recognition tasks, particularly when high-quality audio recordings are available.

In contrast, the MELD dataset shows no improvement when using speech descriptions and only a modest improvement when adding speech impressions, with a slight increase from 67.44\% to 67.604\%. This mirrors the trends observed in the projection-based model experiments, where the lower audio quality in MELD limits the effectiveness of speech feature integration.

Interestingly, the inclusion of speech impressions does not result in further improvements over speech descriptions for either dataset. In IEMOCAP, the performance (71.542\%) with speech impressions falls between the text-only and text with speech descriptions results, while in MELD, it performs similarly to the text-only baseline (67.604\%). This suggests that the hard-coded impressions, though more interpretative, may introduce noise or inaccuracies that the model struggles to utilize effectively. The lack of improvement with speech impressions implies that allowing the model to learn directly from objective speech descriptions might be a more reliable approach.

Additionally, we compare the performance of numerical and categorical acoustic features using simple machine learning models, as detailed in Appendix \ref{sec:appendix_3}. 

\subsubsection{Can LLMs Leverage Speech Features Alone?}

\begin{figure}[t]
  \includegraphics[width=\columnwidth]{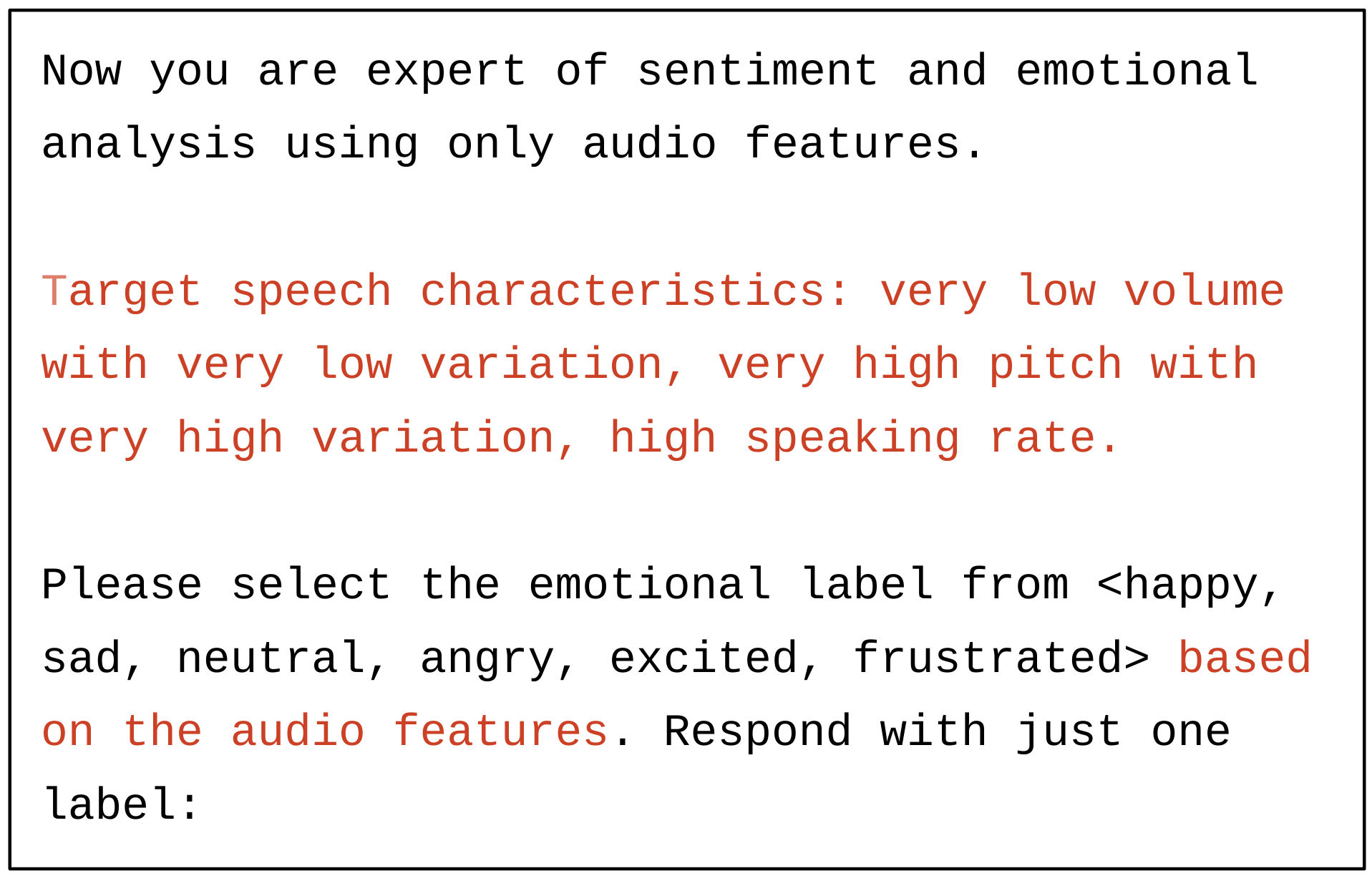}
  \caption{LLM Prompt Template for speech-feature-only Emotion Detection.}
  \label{fig:espeech_temp}
\end{figure}

\begin{table}[ht]
\centering
\begin{tabular}{l|c}
\hline
\textbf{Method}  & \textbf{Weighted F-1} \\
\hline
blind estimation & 16.67 \\
\hline
numerical features (ML) & 32.0\\
\hline
3-class categories (LLM) & 27.602 \\
5-class categories (LLM) & 27.895 \\
speech impression (LLM) & 27.794 \\
\hline
\end{tabular}
\caption{Comparison of Weighted F-1 Scores for Different Classification Methods on IEMOCAP \textbf{using speech-feature only}}
\label{tab:weighted-f1-scores}
\end{table}

\begin{table*}
  \centering
  \begin{tabular}{lccc}
    \hline
    \textbf{Method} & \textbf{Text-only} & \textbf{Text + speech des} & \textbf{Text + speech imp}\\
    \textbf{Model} & &\\
    \hline
    Claude Sonnet 3.5 (zero-shot) & 58.99 & \textbf{59.23} \textcolor{darkgreen}{\scriptsize{(+0.24)}} & 59.11 \textcolor{darkgreen}{\scriptsize{(+0.12)}} \\
    LLaMA-3-8b-instruct (zero-shot) & 35.14 \scriptsize{$\pm$0.16} & 44.18 \scriptsize{$\pm$0.56} \textcolor{darkgreen}{\scriptsize{(+9.04)}} & \textbf{45.05} \scriptsize{$\pm$0.68} \textcolor{darkgreen}{\scriptsize{(+9.91)}} \\ 
    phi-3-13b-instruct (zero-shot) & 41.63 \scriptsize{$\pm$0.06} & 45.85 \scriptsize{$\pm$0.02} \textcolor{darkgreen}{\scriptsize{(+4.22)}} & \textbf{46.95} \scriptsize{$\pm$0.04} \textcolor{darkgreen}{\scriptsize{(+5.32)}} \\ 
    \hline
    LLaMA-2-7b-base (LoRA) & 70.11 \scriptsize{$\pm$0.36} & \textbf{72.02} \scriptsize{$\pm$0.54} \textcolor{darkgreen}{\scriptsize{(+1.91)}} & 71.54 \scriptsize{$\pm$1.12} \textcolor{darkgreen}{\scriptsize{(+1.43)}} \\ 
    LLaMA-3-8b-base (LoRA) & 70.32 \scriptsize{$\pm$0.95} & 71.74 \scriptsize{$\pm$0.79} \textcolor{darkgreen}{\scriptsize{(+1.43)}} & \textbf{71.91} \scriptsize{$\pm$0.62} \textcolor{darkgreen}{\scriptsize{(+1.59)}} \\
    LLaMA-3-8b-instruct (LoRA) & 70.47 \scriptsize{$\pm$0.94} & \textbf{72.1} \scriptsize{$\pm$0.65} \textcolor{darkgreen}{\scriptsize{(+1.63)}} & 71.13 \scriptsize{$\pm$0.36} \textcolor{darkgreen}{\scriptsize{(+0.66)}} \\
    phi-3-13b-instruct (LoRA) & 69.22 \scriptsize{$\pm$0.73} & 70.45 \scriptsize{$\pm$0.86} \textcolor{darkgreen}{\scriptsize{(+1.23)}} & \textbf{70.58} \scriptsize{$\pm$0.24} \textcolor{darkgreen}{\scriptsize{(+1.36)}} \\
    \hline
  \end{tabular}
  \caption{\textit{SpeechCueLLM} performances (weighted F1 score in \%) on IEMOCAP dataset across different models and input modalities. Results for LoRA fine-tuned models are reported as mean $\pm$ standard deviation over three runs. Bold indicates the best performance for each model. Dark green values in parentheses show the improvement over the Text-only method.}
  \label{tab:model_comparison}
  \vspace{-0.3cm}
\end{table*}



To further validate the effectiveness of simple speech features, we conduct experiments using only speech features in the prompt, as shown in Figure \ref{fig:espeech_temp}. The results in Table \ref{tab:weighted-f1-scores} demonstrate the potential of LLMs in leveraging these features for emotion classification, significantly outperforming the blind estimation baseline of 16.67\%. Specifically, the 5-class categorical representation with LLMs achieves a weighted F1 score of 27.895\%, while the 3-class representation yields 27.602\%, suggesting that decreasing the granularity of speech feature categorization does not significantly affect the LLM's ability to classify emotions.

As a reference point, using numerical feature representations with traditional machine learning (ML) models results in a weighted F1 score of 32\%. The lower performance of the speech-only LLM approach compared to the ML approach highlights key differences in how these models process information. ML models, designed for discriminative tasks, can directly exploit structured, one-hot encoded categorical features. In contrast, LLMs interpret natural language descriptions, introducing a layer of abstraction and potential ambiguity. Despite this performance gap, LLMs' performance remains substantially above the blind estimation baseline, indicating that they can still extract meaningful emotional cues from the speech descriptions.

\subsubsection{How Do LLMs Differ in Performance?}
\label{sec:llms}

To evaluate the effectiveness of our approach across different language models, we conduct experiments using several popular open-sourced LLMs, including LLaMA-2, LLaMA-3, and Phi-3, along with a powerful closed-sourced model, Claude Sonnet 3.5. Table \ref{tab:model_comparison} presents the emotion recognition accuracy on the IEMOCAP dataset for these models under different input conditions: text-only, text with speech descriptions, and text with speech impressions.

The substantial performance gap between the zero-shot model and the fine-tuned models highlights the importance of task-specific fine-tuning for emotion recognition. However, even in a zero-shot setting, the models benefit from the inclusion of speech features, further emphasizing the value of multimodal inputs in emotion detection.

We observed significant improvements after applying speech descriptions in the zero-shot setting, with almost a 10-point increase in F1 scores for LLaMA-3-8b. This signifies the effectiveness of SpeechCueLLM and expands the use cases for applications that aim for fast deployment or do not have the resources for fine-tuning. The speech impressions also consistently improve upon the speech descriptions in the zero-shot setting for both models.

In the LoRA setting, our results show consistent performance improvements across all models when either speech descriptions or impressions are integrated, underscoring the value of audio-derived features in emotion recognition tasks. In most cases, speech descriptions slightly outperform speech impressions, though the differences are often marginal. This suggests that more objective, feature-based descriptions may be more reliable than hard-coded speech impressions for emotion detection tasks.

Interestingly, increasing model size does not necessarily lead to better performance in this task. Despite their stronger performance on public benchmarks \cite{abdin2024phi3technicalreporthighly,dubey2024llama3herdmodels}, larger models like LLaMA-3 and Phi-3 do not consistently outperform the smaller LLaMA-2 model when fine-tuned for emotion detection. For instance, Phi-3, the largest model tested (13B parameters), performs worse than both LLaMA-2 and LLaMA-3 across all input conditions. Although LLaMA-3-8b-instruct achieves the highest accuracy (72.098\%) when provided with speech descriptions, the improvement over LLaMA-2-7b-base (72.021\%) is minimal. The performance differences between the instruct and base versions of these models are insignificant for this task.

\subsubsection{Why SpeechCueLLM Gets Better Scores?}

As highlighted in Table \ref{tab: emotions}, our analysis across three independent runs demonstrates consistent improvements in the F1-score for all emotion classes when leveraging SpeechCueLLM. This underscores the robustness and reliability of the SpeechCueLLM approach in enhancing emotion recognition across diverse emotional categories.

The confusion matrices in Table \ref{tab:confuse1} and Table \ref{tab:confuse2} reveal a notable improvement in the accurate classification of 'Neutral' emotions, increasing from 74.80\% without speech features to 79.81\% with SpeechCueLLM. While other emotion classes exhibit incremental gains, this substantial improvement in 'Neutral' classification highlights SpeechCueLLM's ability to discern the subtle nuances of emotionally subdued or calm utterances. This suggests that the model effectively captures the distinct characteristics of neutral expressions, particularly in scenarios where speakers convey emotions with minimal intensity or through a calm demeanor.





\section{Conclusions}

In summary, our contributions are twofold: first, we introduce \textit{SpeechCueLLM}, a novel approach that translates audio features into natural language descriptions to enable multimodal emotion analysis without architectural changes. This method significantly improves performance, especially with high-quality audio, and outperforms baseline models that rely on structural modifications.

Second, our key findings highlight the effectiveness and limitations of the method, emphasizing the importance of audio quality and the benefits of prompt-based techniques. Incorporating speech descriptions improves emotion recognition across LLMs, though the approach’s success depends on audio quality, as shown by varying dataset results. Our experiments also demonstrate that prompt-based methods outperform direct integration of speech encoder features, leveraging LLMs’ strengths in language processing.



Future work could prioritize on improving feature extraction techniques to better handle noisy, real-world audio data, ensuring robustness across diverse audio conditions. Moreover, advancing natural language representations of speech characteristics could further enhance both interpretability and accuracy in multimodal emotion recognition.
Expanding this methodology to other domains and modalities also presents exciting opportunities for further research.


\clearpage

\section*{Limitations}
While our approach to enhancing LLM-based emotion detection shows promise, it is important to recognize its limitations. A major constraint is the method's reliance on audio quality. As shown by the contrasting results between the IEMOCAP and MELD datasets, performance degrades significantly when the input audio is noisy or of low quality.

Another challenge is the nature of the training datasets. Our study primarily relies on acted (IEMOCAP) or scripted (MELD) emotional expressions, which may not fully capture the complexity and subtlety of emotions in natural, spontaneous speech. This dataset bias could limit the generalizability of our findings to real-world scenarios, where emotions are often more nuanced, mixed, or ambiguously expressed. The difference between performed emotions in our datasets and the diverse emotional expressions in everyday life may impact the system's effectiveness in real-world applications.

Finally, the computational resources required for this approach present a practical limitation. Fine-tuning large language models is computationally intensive, which may hinder the implementation of this method in resource-constrained environments. This requirement could restrict the accessibility and widespread adoption of our approach, especially in settings where high-performance computing resources are not readily available.

Addressing these limitations in future research will be critical for developing more robust, versatile, and scalable emotion recognition systems that can handle the complexities of real-world emotional expressions across diverse audio conditions and computational environments.


\bibliography{anthology,custom}
\clearpage

\appendix

\section{Appendix}
\label{sec:appendix}

\subsection{Contextual Speech Features}
\label{sec:appendix_2}

\begin{table}[ht]
\centering
\begin{tabular}{l|c}
\hline
\textbf{Method} (IEMOCAP) & \textbf{Weighted F-1} \\
\hline
text-only & 70.111 \scriptsize{$\pm$0.36} \\
+ speech des & 72.021 \scriptsize{$\pm$0.54} \\
\hline
+ volume (all utterances) & 71.226 \scriptsize{$\pm$0.25} \\
+ volume (the last three) & 71.487 \scriptsize{$\pm$0.59} \\
+ pitch (all utterances) & 71.28 \scriptsize{$\pm$0.47}\\
+ pitch (the last three) & \textbf{72.595} \scriptsize{$\pm$0.26} \\
+ all features (the last three) & 71.461 \scriptsize{$\pm$0.76} \\
\hline
\end{tabular}
\caption{Comparison of Weighted F-1 Scores for Different Methods of Adding Speech-Feature Context on IEMOCAP. All experiments in the table use LoRA on the LLaMA-2-7b-base model.}
\label{tab:speech_context}
\end{table}


In our exploration of incorporating speech features into context utterances, we found that selective application of these features yielded better results than a blanket approach. As shown in Table \ref{tab:speech_context}, adding speech descriptions to only the last three utterances in the context proved more effective than applying them to all utterances. This suggests that recent context plays a more significant role in emotion recognition. Notably, pitch features outperformed volume features, with pitch information from the last three utterances providing a 0.6\% improvement over non-contextual speech descriptions, achieving a weighted F1 score of 72.595\%.

Interestingly, combining all speech features for the context utterances did not enhance performance, likely due to information overload. These findings underscore the importance of selective feature inclusion and careful consideration of context in multimodal emotion recognition tasks, where focusing on recent and relevant speech features can improve results without overwhelming the model with excess information.

\subsection{Evaluation of Speech Features Using Simple Machine Learning Models}
\label{sec:appendix_3}

\begin{table*}
\centering
\begin{tabular}{lccc|c}
\hline
\textbf{Features} & RF & SVM & MLP & \textbf{Average} \\
\hline
Random Guess & -- & -- & -- & 0.167(0.167) \\
\hline
Numerical & 0.33(0.31) & 0.33(0.31) & 0.30(0.29) & \textbf{0.320}(\textbf{0.303}) \\
\hline
3-class & 0.31(0.28) & 0.32(0.30) & 0.30(0.27) & 0.310(0.283) \\
4-class & 0.28(0.27) & 0.32(0.29) & 0.29(0.27) & 0.297(0.277) \\
5-class & 0.31(0.29) & 0.33(0.30) & 0.30(0.29) & 0.313(0.293) \\
6-class & 0.30(0.28) & 0.31(0.29) & 0.29(0.28) & 0.300(0.283) \\
\hline
\end{tabular}
\caption{Performance comparison of discriminant ML models using different feature representations on IEMOCAP using weighted F1(Macro F1) scores}
\label{tab:iemocap_audio_feature_modeling}
\end{table*}

\begin{table*}
\centering
\begin{tabular}{lccc|c}
\hline
\textbf{Features} & RF & SVM & MLP & \textbf{Average} \\
\hline
Random Guess & -- & -- & -- & 0.143(0.143) \\
\hline
Numerical & 0.37(0.15) & 0.34(0.12) & 0.38(0.18) & 0.363(0.150) \\
\hline
3-class & 0.37(0.15) & 0.36(0.13) & 0.37(0.16) & \textbf{0.367}(0.147) \\
4-class & 0.36(0.15) & 0.35(0.13) & 0.36(0.14) & 0.357(0.140) \\
5-class & 0.36(0.16) & 0.35(0.13) & 0.36(0.15) & 0.357(0.147) \\
6-class & 0.36(0.17) & 0.35(0.13) & 0.35(0.17) & 0.353(\textbf{0.157}) \\
\hline
\end{tabular}
\caption{Performance comparison of discriminant ML models using different feature representations on MELD using weighted F1(Macro F1) scores}
\label{tab:meld_audio_feature_modeling}
\end{table*}

\begin{table*}
    \centering
    \begin{tabular}{lcc}
        \hline
        Emotion     & F1-score (Without Speech Features) & F1-score (SpeechCueLLM) \\
        \hline
        Happy       & 59.73\%                            & \textbf{60.56\%} (+0.83\%)       \\
        Sad         & 80.23\%                            & \textbf{82.21\%} (+1.98\%)       \\
        Neutral     & 71.84\%                            & \textbf{73.75\%} (+1.91\%)       \\
        Angry       & 65.54\%                            & \textbf{66.41\%} (+0.87\%)       \\
        Excited     & 71.55\%                            & \textbf{73.71\%} (+2.16\%)       \\
        Frustrated  & 68.25\%                            & \textbf{69.56\%} (+1.31\%)       \\
        \hline
    \end{tabular}
    \caption{F1-Scores for Different Emotions}
    \label{tab: emotions}
\end{table*}

\begin{table*}
    \centering
    \begin{tabular}{lcccc}
        \hline
        Actual $\backslash$ Predicted & angry+frustracted & happy\_exc & neutral & sad \\
        \hline
        angry+frustracted & \textbf{86.02\%} & 0.36\% & 9.25\% & 4.35\% \\
        happy\_exc        & 0.75\%           & \textbf{83.77\%} & 13.51\% & 1.96\% \\
        neutral           & 14.86\%          & 6.34\%  & \textbf{74.80\%} & 4.00\% \\
        sad               & 12.24\%          & 0.41\%  & 7.21\% & \textbf{80.14\%} \\
        \hline
    \end{tabular}
    \caption{Normalized Confusion Matrix - Without Speech Features}
    \label{tab:confuse1}
\end{table*}

\begin{table*}
    \centering
    \begin{tabular}{lcccc}
        \hline
        Actual $\backslash$ Predicted & angry+frustracted & happy\_exc & neutral & sad \\
        \hline
        angry+frustracted & \textbf{86.21\%} & 0.73\% & 10.16\% & 2.90\% \\
        happy\_exc        & 1.20\%           & \textbf{82.52\%} & 14.69\% & 1.58\% \\
        neutral           & 10.79\%          & 5.74\%  & \textbf{79.81\%} & 3.57\% \\
        sad               & 10.76\%          & 0.68\%  & 8.17\% & \textbf{80.38\%} \\
        \hline
    \end{tabular}
    \caption{Normalized Confusion Matrix - SpeechCueLLM}
    \label{tab:confuse2}
\end{table*}

To evaluate the effectiveness of our extracted speech features in capturing emotion-related information, we conducted experiments using various machine learning models on both numerical and categorical representations of these features. Tables \ref{tab:iemocap_audio_feature_modeling} and \ref{tab:meld_audio_feature_modeling} present the performance of Random Forest (RF), Support Vector Machine (SVM), and Multi-Layer Perceptron (MLP) models on the IEMOCAP and MELD datasets.

Our analysis of the IEMOCAP dataset reveals that the five extracted speech features contain substantial emotion-related information. Both weighted F1 and macro F1 scores (approximately 0.32 and 0.30, respectively) are nearly double the random guess baseline of 0.167. This significant improvement over the baseline indicates that the speech features are effectively capturing emotional cues. Notably, converting numerical features into categorical features does not result in a significant loss of emotion-related information. Additionally, dividing the numerical features into five classes yields the best performance, with weighted F1 and macro F1 scores of 0.313 and 0.293, respectively, suggesting that a moderate level of discretization preserves clear emotional patterns in speech features.

The results from the MELD dataset highlight a crucial aspect of speech-based emotion recognition: the importance of audio quality. While weighted F1 scores (around 0.36) surpass the random guess baseline of 0.143, the macro F1 scores (approximately 0.15) are only marginally better than the baseline. This discrepancy is important: MELD is highly imbalanced, with neutral emotions comprising about 50\% of the dataset. The high weighted F1 score primarily reflects the model’s ability to predict the dominant neutral class, rather than its capacity to distinguish between other emotions from speech features. In contrast, the near-baseline macro F1 score, which treats all classes equally, reveals that the extracted speech features provide minimal informative value across diverse emotion categories.

This lack of substantial improvement in macro F1 scores indicates that the low quality and challenging nature of the audio in MELD are key factors in the poor performance. These findings underscore the critical importance of audio quality in obtaining reliable speech features for emotion recognition tasks. They also explain the modest gains observed when incorporating speech descriptions in our earlier language model experiments on MELD, in contrast with the significant improvements seen in IEMOCAP.

\subsection{Additional Speech Features Processing Tables}

\label{sec:appendix_1}
\begin{table*}[ht]
\centering
\begin{tabular}{|c|l|}
\hline
\textbf{Number of Classes} & \textbf{Quantile Splits} \\
\hline
3 & [0.25, 0.75] \\
\hline
4 & [0.25, 0.5, 0.75] \\
\hline
5 & [0.1, 0.25, 0.75, 0.9] \\
\hline
6 & [0.1, 0.25, 0.5, 0.75, 0.9] \\
\hline
\end{tabular}

\caption{Quantile splits for different numbers of classes}
\label{table:quantile_splits}
\end{table*}

\begin{table*}[ht]
\centering
\resizebox{\textwidth}{!}{%
\begin{tabular}{|l|l|l|}
\hline
\textbf{Feature} & \textbf{Level} & \textbf{Impression} \\ \hline
Pitch & High/Very High & Uses a higher pitch \\ \cline{2-3} 
 & Low/Very Low & Uses a lower pitch \\ \cline{2-3} 
 & Medium & Has a moderate pitch \\ \hline
Pitch Variation & High/Very High & With noticeable variation, suggesting expressiveness \\ \cline{2-3} 
 & Low/Very Low & That remains steady, potentially indicating calmness or seriousness \\ \cline{2-3} 
 & Medium & With typical variation \\ \hline
Volume & High/Very High & Speaking loudly, which might indicate excitement, confidence, or urgency \\ \cline{2-3} 
 & Low/Very Low & Speaking softly, possibly suggesting calmness, shyness, or caution \\ \cline{2-3} 
 & Medium & Using a moderate volume \\ \hline
Volume Variation & High/Very High & With significant volume changes \\ \cline{2-3} 
 & Low/Very Low & With little volume variation \\ \cline{2-3} 
 & Medium & With normal volume variation \\ \hline
Speech Rate & High/Very High & Talking quickly, which could indicate excitement, urgency, or nervousness \\ \cline{2-3} 
 & Low/Very Low & Talking slowly, possibly suggesting thoughtfulness, hesitation, or calmness \\ \cline{2-3} 
 & Medium & Speaking at a moderate pace \\ \hline
\end{tabular}
}
\caption{Feature-Impression Mapping}
\label{table:Feature-Impression Mapping}

\end{table*}

\subsection{Baseline Model Figure}
\begin{figure*}[t]
\centering
  \includegraphics[width=\columnwidth]{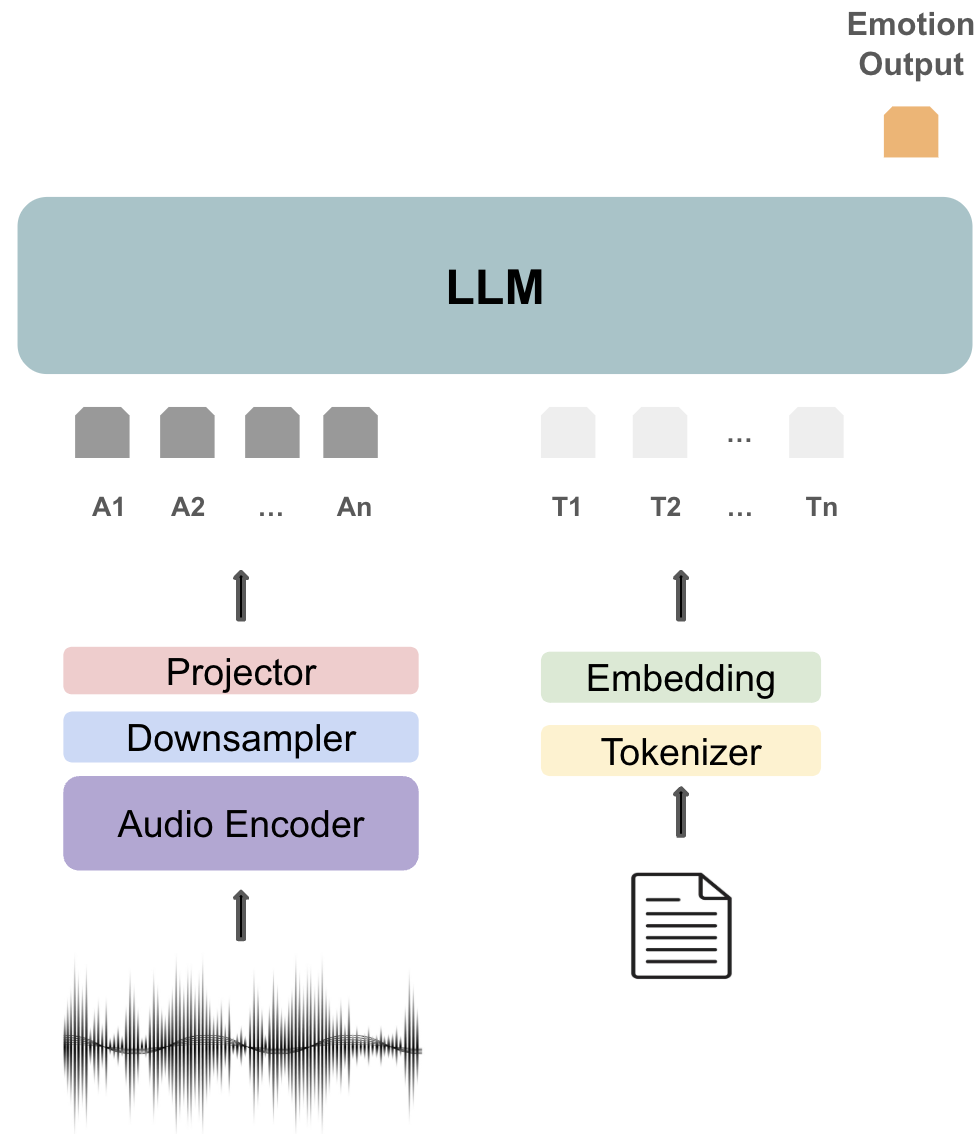}
  \centering
  \caption{The baseline model structure involves projecting speech encoder features directly into the LLM embedding space.}
  \label{fig:projector}
\end{figure*}

\end{document}